\documentclass[letterpaper, 10 pt, conference]{ieeeconf}  % Comment this line out if you need a4paper

\IEEEoverridecommandlockouts                              % This command is only needed if
\usepackage{amsthm}
\usepackage{amssymb}  % assumes amsmath package installed

\usepackage{graphics} % for pdf, bitmapped graphics files
\usepackage{float,url,enumerate}
\usepackage{graphicx}
\usepackage{wrapfig}
\usepackage{hyperref}

\usepackage[ruled,vlined,linesnumbered]{algorithm2e}

\usepackage{multirow} % for tables
\usepackage{array} % for setting width of a table

\usepackage[usenames, dvipsnames]{color} % for comments

\usepackage{xspace}
\makeatletter
\DeclareRobustCommand\onedot{\futurelet\@let@token\@onedot}
\def\@onedot{\ifx\@let@token.\else.\null\fi\xspace}
 
\def\ie{\emph{i.e}\onedot} 
 
\def\etc{\emph{etc}\onedot}

\makeatother

\newcommand{\C}{\mathcal{C}} % special contact frame

\newcommand{\G}{{\bf G}} % set of grasp points
\newcommand{\cond}{{\rm cond}} % set of vertices

\title{\vspace{-0cm}
\LARGE \bf
An Efficient Closed-Form Method for Optimal Hybrid Force-Velocity Control
}

\author{Yifan Hou and Matthew T. Mason~\IEEEmembership{Fellow,~IEEE} % <-this % stops a space
\thanks{*This work was supported under NSF Grant IIS-1813920 and IIS-1909021. }% <-this % stops a space
\thanks{The authors are with the Robotics Institute, Carnegie Mellon University, Pittsburgh, PA 15213, USA.
        {\tt\small yifanh@cmu.edu, mattmason@cmu.edu}}
}

\begin{document}

\maketitle
\thispagestyle{empty}
\pagestyle{empty}

%%%%%%%%%%%%%%%%%%%%%%%%%%%%%%%%%%%%%%%%%%%%%%%%%%%%%%%%%%%%%%%%%%%%%%%%%%%%%%%%
\begin{abstract}
This paper derives a closed-form method for computing hybrid force-velocity control. The key idea is to maximize the kinematic conditioning of the mechanical system, which includes a robot, free objects, a rigid environment and contact constraints.
The method is complete, in that it always produces an optimal/near optimal solution when a solution exists.
It is efficient, since it is in closed form, avoiding the iterative search of our previous work.
We test the method on 78,000 randomly generated test cases. The method  outperforms our previous search-based technique by being from 7 to 40 times faster, while consistently producing better solutions in the sense of robustness to kinematic singularity.
We also test the method in several representative manipulation experiments.

% Contacts introduce challenges to reliably execution of rigid body manipulation.
% Unexpected slipping and sticking can drastically change the force balance in the system and cause the task to fail.
% Violation of contact constraints will bring huge internal forces that break the rigid body modeling.
% %
% Our previous work \cite{HouICRA19Hybrid} has shown that a well-designed Hybrid Force-Velocity Control (HFVC) can effectively avoid these undesired situations, even under modeling uncertainties and force disturbances. The task of finding the most robust HFVC for a given manipulation problem is called \textit{Hybrid Servoing (HS)}.
% %
% The original hybrid servoing algorithm used an approximated robustness criteria and formulated a non-convex optimization problem, which requires a long time to solve and does not guarantee the quality of solution. The result is a slow and sometimes choppy robot motion.
% % todo: talk about experimental results
% In this paper, we propose \textit{Optimally-Conditioned Hybrid Servoing (OCHS)} to solve the hybrid servoing problem without approximation. OCHS uses matrix decomposition instead of non-convex optimization, 
% which results in a 7x to 40x speedup and guaranteed optimality. The latter is crucial to contact-critical tasks where one bad action at any time step can cause a failure.
% %
% We compare the performance of OCHS with the original hybrid servoing algorithm in extensive simulation tests as well as representative experiments.
% Link to the video: \href{https://youtu.be/KtSNmvwOenM}{https://youtu.be/KtSNmvwOenM}.

\end{abstract}

%%%%%%%%%%%%%%%%%%%%%%%%%%%%%%%%%%%%%%%%%%%%%%%%%%%%%%%%%%%%%%%%%%%%%%%%%%%%%%%%
%%%%%%%%%%%%%%%%%%%%%%%%%%%%%%%%%%%%%%%%%%%%%%%%%%%%%%%%%%%%%%%%%%%%%%%%%%%%%%%%

% !TEX root = ../ICRA21Analytical.tex
\section{INTRODUCTION}
\label{sec:intro}
Contact constraints help human manipulation with improved precision and dexterity. For example, when cutting a piece of wood with  a band saw, we often slide the wood along a guide rail to position it accurately.
% Contact constraints also improves human manipulation dexterity in some tasks, such as electronics assembly where only one fingertip is allowed in the confined space.
% However, the contact constraints make those tasks difficult for robotic automation.
% Traditional high-stiffness industrial robots may cause huge internal force or break contacts unexpectedly if the robot action deviates from the constraints. Low-stiffness, ``collaborative'' robots can respond to contact forces to avoid extensive forces, however, they are also equally sensitive to disturbance forces.
%
Hybrid Force-Velocity Control (HFVC) naturally suits such tasks. The velocity control (high-stiffness) can drive the system precisely, avoiding the need to finely balance the forces. The force control (low-stiffness) can avoid excessive internal force and maintain desired contacts. A properly designed HFVC could keep both advantages.

Under modeling uncertainties, HFVC may become infeasible, violating constraints and generating huge forces.
We describe this phenomenon by treating the robot-object-environment system as a kinematic chain connected by contact constraints, then use its kinematic conditioning to evaluate the quality of the HFVC. A well conditioned system can remain feasible under large modeling errors.
Our goal in this work is to compute a HFVC that maximizes the kinematic conditioning of the manipulation system.

While kinematic conditioning of a fully-actuated manipulator was well studied \cite{angeles1992kinematic}\cite{angeles1992design}\cite{salisbury1982articulated}, kinematic conditioning of a system with free objects still needs a clear characterization \cite{Hou2019Criteria}. Our previous work \cite{HouICRA19Hybrid} approximated the condition number of the whole system with a polynomial and optimized it in a non-convex optimization. Given enough initial guesses and sufficient number of iterations, the algorithm could find a good solution. However, the trade-off between computation speed and quality of solution is limiting its applications, especially in industrial applications where both speed and safety guarantee are required.

% With abundant HFVC implementations available \cite{salisbury1980active,raibert1981hybrid,hogan1985impedance,khatib1987unified,Maples1986Experiments,lopes2008force}, this paper focus on the remaining challenge: computing the ``best'' HFVC to execute a given motion plan, which includes computing the dimensionality, direction and magnitude of both force and velocity control at every time step. We called this problem \textit{Hybrid Servoing (HS)} in \cite{HouICRA19Hybrid}.

This paper aims to solve the speed and optimality problem. We make three contributions: first, we provide a precise characterization of the conditioning of a robot-object-environment system. Second, we present \textit{Optimally-Conditioned Hybrid Servoing (OCHS)}, an algorithm that efficiently computes well-conditioned HFVC. OCHS avoids non-convex optimization, which leads to a 7 to 40 times speed up comparing with our previous work. Third, we test our algorithm extensively in randomly generated problems as well as experiments to demonstrate its computation speed and quality consistency.

% Hybrid servoing can be used as a tracking controller to reliably execute pre-computed motion plans with sticking contacts. When sliding contacts are present, we use hybrid servoing in the more elaborate framework, wrench stamping \cite{Hou2020SharedGrasping}, to evaluate stability and compute robust control.

The paper is organized as follows. In the next section we review the related work. In Section~\ref{sec:modeling} we introduce the modeling for a manipulation problem under contact constraints. Then we introduce the problem formulation in Section~\ref{sec:hybrid_servoing_problem}. Next, we derive the OCHS algorithm in Section~\ref{sec:approach}. Section~\ref{sec:evaluation} evaluates the algorithm in randomly generated test problems, while multiple representative experimental results are presented in Section~\ref{sec:experiments}.

% !TEX root = ../ICRA21Analytical.tex
\section{RELATED WORK}
\label{sec:related_work}

\subsection{Manipulation with Hybrid Force-Velocity Control}
% It is suitable for executing manipulation under contact constraints:
% the velocity control provides disturbance rejection ability and can be used for
% executing desired velocity robustly, while the force control can accommodate modeling errors and maintain desired contacts.
The idea of HFVC was originally introduced in \cite{mason1981compliance}, which provides a framework for identifying force and velocity-controlled directions in a task frame given a task description. The framework was then completed and implemented in \cite{raibert1981hybrid}. For the control of a manipulator subject to constraints, it was common to align the force and velocity control directions with the row and null space of the contact Jacobian \cite{west1985method}\cite{yoshikawa1987dynamic}. The approach has industrial applications including polishing \cite{nagata2007cad} and peg-in-hole assembly \cite{park2017compliance}.

However, when the system contains one or more free objects with no attachment to any motor, most previous work was case by case study, such as \cite{dehio2018modeling} and \cite{uchiyama1988symmetric}. In some special cases, it is possible to design a HFVC from simple heuristics using local contact information, such as in multi-finger grasping \cite{murray1994mls} and locomotion \cite{buschmann2009biped}\cite{fujimoto1996proposal}. Our previous work \cite{HouICRA19Hybrid} proposed the first general algorithm for manipulation under rigid contacts using HFVC.

Once we have a HFVC, the final step is to implement it on a robot. HFVC can be implemented by wrapping around force control \cite{hogan1985impedance}\cite{khatib1987unified}\cite{raibert1981hybrid}\cite{salisbury1980active} or position/velocity control \cite{lopes2008force}\cite{Maples1986Experiments}. The choice depends on the type of robot. We refer the readers to Whitney \cite{whitney1987historical} and De Schutter \cite{de1998force} for comparisons of different HFVC implementations.

\subsection{Robustness of Manipulation System under HFVC} % (fold)
\label{sub:robustness_of_manipulation_under_hfvc}
There is plenty of work on the stability of a hybrid force-velocity controlled system under active contact with the environment. Lagrange dynamics modeling and Lyapunov stability analysis have been conducted on the whole constrained robot system \cite{brogliato1997control}\cite{eppinger1987introduction}\cite{kazerooni1990stability}\cite{mcclamroch1988feedback}.
Much attention was paid to the stability of the engaging/disengaging process \cite{ishikawa1989stable}\cite{mills1993control}\cite{volpe1993theoretical}, which is particularly difficult to stabilize.

Most of these analyses focused on the stability of the inner control loop, i.e. whether the robot action would converge, oscillate, or diverge. However, a stable robot may still fail a manipulation task for two reasons. The first is ill-conditioning under modeling errors. The second is unexpected changes of contact modes such as unexpected slipping or sticking, which may directly cause task failure or lead to crashing. This work assumes the robot has stable HFVC and focuses on handling the above two failure cases.

 % When the condition number reach the lowest possible value, one, the kinematic system is referred to as \textit{isotropic} \cite{angeles1992kinematic}.
% 
% !TEX root = ../ICRA21Analytical.tex

\section{MODELING}
\label{sec:modeling}
First we introduce the instantaneous modeling of a manipulation system subject to contact constraints, which are mostly consistent with our previous work \cite{HouICRA19Hybrid} except for notational simplifications. Consider a robot and at least one object in a rigid environment. The robot, object(s), and the environment has $n_a$, $n_u$, and zero degree-of-freedoms (DOF), respectively. The total DOF of the system is $n = n_a + n_u$. The subscript `a' and `u' means `actuated' and `unactuated'.
Denote $v=[v^T_u\  v^T_a]^T, f=[f^T_u\ f^T_a]^T \in \mathbb{R}^{n}$ as the generalized velocity and force vectors, where $f_u$ is always zero.
The following assumptions are made:
\begin{itemize}
    \item Motions are quasi-static, \ie inertia force and Coriolis force are negligible.
    \item Object, robot and environment are all rigid.
    \item Friction follows Coulomb's Law.
\end{itemize}

% subsection symbols (end)
\subsection{Contact Constraints on Velocity} % (fold)
\label{sub:contact_modeling_and_constraints}
We consider point contact with clearly defined contact point location and contact normal. This is the case for point-to-face contacts. Edge-to-edge, edge-to-face, and face-to-face contacts can be approximated by one or more point contacts \cite{prehensile}. Point-to-point and point-to-edge contacts are not considered here due to their rare appearance. We consider three types of contact modes: sticking, sliding, and separation. Both sticking and sliding contacts impose a linear constraint in the contact normal direction; a sticking contact also impose constraints in the contact tangential directions. We consider holonomic constraints imposed by the contacts, which are bilateral constraints on the system configuration that are also independent of the system velocity. They are linear constraints on the system velocity:
\begin{equation}
\label{eq: velocity constraint}
    Jv=0,
\end{equation}
where $J$ is the contact Jacobian \cite{murray1994mls}. Equation~(\ref{eq: velocity constraint}) can also model any other holonomic constraints, such as the connection constraint between two links of a robot joint.

\subsection{Goal Description} % (fold)
\label{sub:goal_description}
Users shall provide the control goal, which is an expected system velocity. The goal at a time instant can be written as an affine constraint on the generalized velocity:
\begin{equation}
\label{eq:goal}
    Gv=b_G,
\end{equation}
which can be derived from a given trajectory by taking first-order derivative.
We can use six rows to specify the desired velocity of a rigid body in 3D, or only use three rows to specify a desired rotational velocity.

The goal specification (\ref{eq:goal}) must not be redundant. For example, to slide an object on a planar surface in 3D, the goal should have no more than three rows. It should not specify the object velocity in the contact normal direction, which is already limited to zero by the contact constraint.

\subsection{Constraints on force} % (fold)
\label{subsub:newton_s_second_law}
Denote $\lambda$ as the vector of contact forces. Using the principle of virtual work \cite{villani2008force}, we can write the contribution of $\lambda$ to the generalized force space as $\tau=J'^T\lambda$. Note the $J'$ here is different from the $J$ in (\ref{eq: velocity constraint}), because $J'$ may have more rows that corresponds to sliding friction.

There are two kinds of force constraints. One is the Newton's Second Law under quasi-static approximation:
\begin{equation}
\label{eq:newton_second_law}
     J'^T\lambda + f + F = 0.
\end{equation}
The three terms are contact forces, control actions (internal forces) and external forces, respectively. The external force $F \in \mathbb{R}^{n}$ may include gravity, disturbance forces, \etc.

The other force constraint is the condition for staying in the desired contact mode, we called them the \textit{guard conditions} \cite{HouICRA19Hybrid}. It's usually a good practice to make the guard condition stricter than necessary to encourage conservative actions. We consider guard conditions that are affine constraints on force variables. Examples are friction cone constraints and lower/upper bounds on forces.
\begin{equation}
\label{eq:guard_conditions}
    \Lambda\left[ {\begin{array}{*{20}{c}}
    \lambda\\
    f
    \end{array}} \right] \le {b_\Lambda}.
\end{equation}
Note that (\ref{eq:guard_conditions}) has no equality constraints, so we do not consider sliding friction. This is because applying force on the friction cone is not a robust way to execute a sliding contact \cite{Hou2019Criteria}. A more reliable approach requires model information beyond the scope of this work \cite{Hou2020SharedGrasping}.

\subsection{Hybrid Force-Velocity Control} % (fold)
\label{sub:hybrid_force_velocity_control}
Consider a HFVC with $n_{av}$ dimensions of velocity control and $n_{af}$ dimension of force control, $n_{av} + n_{af} = n_a$.
We use matrix $T\in \mathbb{R}^{n\times n}$ to describe the directions of force/velocity control. Matrix $T$ is diagonal: $T=diag(I_u, R_a)$, where $I_u\in \mathbb{R}^{n_u\times n_u}$ is an identity matrix, $R_a\in \mathbb{R}^{n_a\times n_a}$ is an unitary matrix describing the control axes. Here we assume $R_a$ is orthonormal, so that the force and velocity controls are reciprocal. Without loss of generality, we assume the last $n_{av}$ rows of $T$ are velocity-controlled directions, preceded by $n_{af}$ rows of force-controlled directions.
Denote $w = Tv, \eta=Tf \in\mathbb{R}^{n}$ as the \textit{transformed generalized velocity} and the \textit{transformed generalized force}. We know $w=[w^T_u\ w^T_{af}\ w^T_{av}]^T$, where $w_u=v_u$ is the unactuated velocity, $w_{af}\in \mathbb{R}^{n_{af}}$ is the velocity in the force-controlled directions, $w_{av}\in\mathbb{R}^{n_{av}}$ is the velocity control magnitude. Similarly, $\eta=[\eta^T_u\ \eta^T_{af}\ \eta^T_{av}]^T$, where $\eta_u=f_u=0$ is the unactuated force, $\eta_{af}\in \mathbb{R}^{n_{af}}$ is the force control magnitude, $\eta_{av}\in\mathbb{R}^{n_{av}}$ is the force in the velocity-controlled directions.

To fully describe a HFVC, we need to solve for $n_{av}, n_{af}, R_a, w_{av}$ and $\eta_{af}$.

% !TEX root = ../ICRA21Analytical.tex

\section{THE HYBRID SERVOING PROBLEM}
\label{sec:hybrid_servoing_problem}
Hybrid servoing \cite{HouICRA19Hybrid} is the problem of computing the best HFVC for a constrained manipulation. In this section we derive its cost function from kinematic conditioning then introduce the hybrid servoing problem.
\subsection{Kinematic Conditioning of Manipulation System} % (fold)
\begin{figure}[ht]
    \centering
    \includegraphics[width=0.4\textwidth]{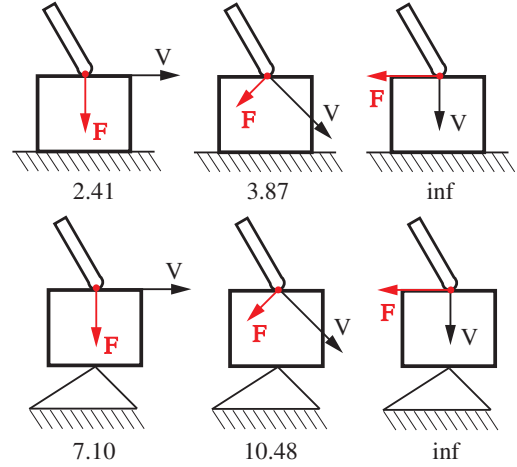}
    \caption{2D examples of HFVC and their corresponding crashing indexes. The robot execute 2D HFVC, with 1D force control and 1D velocity control.}
    \label{fig:crashing_index_examples}
\end{figure}

\label{sub:conditioning_velocity_constraints}
In manipulator kinematic analysis, it is well-known that the condition number of the manipulator Jacobian is an indicator of the kinematic performance of the system \cite{angeles1992kinematic}\cite{angeles1992design}\cite{salisbury1982articulated}. In a manipulation problem with free objects, the kinematic constraints includes (\ref{eq:velocity control}) and the velocity control of HFVC:
\begin{equation}
    \label{eq:velocity control}
    Cv = b_c,
\end{equation}
where $C$ is the last $n_{av}$ rows of $T$, $b_C$ is simply $w_{av}$.
The combined kinematic system is:
\begin{equation}
    \label{eq: combined velocity constraints}
    \left[\begin{array}{c}J\\C\end{array}\right]v = \left[\begin{array}{c}0\\b_C\end{array}\right].
\end{equation}
The condition number of its coefficient matrix needs to be minimized:
\begin{equation}
    \min_{J,C} \cond(\left[\begin{array}{c}J\\C\end{array}\right]).
\end{equation}
Throughout this paper, we use the 2-norm condition number, defined as the following for any matrix $A$:
\begin{equation}
    \label{eq: condition number - singular value}
    \cond(A) = \|A\|_2\|A^\dagger\|_2 = \frac{\sigma_{\max}(A)}{\sigma_{\min}(A)},
\end{equation}
which is the ratio between the maximum and minimum singular values.
However, for our system, directly computing the above condition number makes little sense for two reasons. First, we only want to evaluate the influence of control $C$. The other part of our coefficient matrix, $J$, is constant and could already be ill-conditioned if the contact modeling is redundant. To singulate the influence of $C$, we replace $J$ with an orthogonal basis of its rows, so it represents the same constraint as (\ref{eq: velocity constraint}) but has a condition number of one. Second, the row scaling of $C$ should not affect our criteria, since scaling both sides of (\ref{eq:velocity control}) does not change our control. However, it does change the condition number. This problem is called \textit{artificial ill-condition}\cite{Stewart1987Collinearity}, the typical solution is to pre-normalize each row of our coefficient matrix. Thus our final expression of kinematic conditioning is:
\begin{equation}
    \label{eq:condition number}
    \min_{C} \cond(\left[\begin{array}{c}\hat J\\\hat C\end{array}\right]),
\end{equation}
where rows of $\hat J$ form an orthonormal basis of rows in $J$; $\hat C$ is $C$ with each row normalized.
Figure~\ref{fig:crashing_index_examples} shows the condition number value of several planar examples. When the control is collinear with constraints, the condition number grows to infinity and a tiny motion can cause huge internal force. We have been calling this situation \textit{crashing} in our previous work, and introduced a ``crashing-avoidance score'' in \cite{Hou2019Criteria} to evaluating it. However, equation~(\ref{eq:condition number}) is a more precise description, we call the cost function the \textit{crashing index}.
\subsection{Problem Formulation} % (fold)
\label{sub:problem_formulation}
The task of hybrid servoing is to solve for:
\begin{enumerate}
    \item the dimensions of force-controlled actions and velocity-controlled actions, $n_{af}$ and $n_{av}$, and
    \item the directions to do force control and velocity control, described by the matrix $T$, and
    \item the magnitude of force/velocity actions: $\eta_{af}$ and $w_{av}$,
\end{enumerate}
so as to minimize the crashing index (\ref{eq:condition number}) subject to the following constraints:
\begin{itemize}
    \item Any $v$ under the robot action shall satisfy the goal constraint (\ref{eq:goal});
    \item Any $f$ under the robot action shall satisfy the guard conditions (\ref{eq:guard_conditions}).
\end{itemize}
We use the word `any' because a HFVC usually cannot uniquely determine $v$ and $f$.

% !TEX root = ../ICRA21Analytical.tex

\section{APPROACH}
\label{sec:approach}
% Although there are many variables in a HFVC, we do not need to solve all of them at once. We can decompose hybrid servoing into several smaller problems, using the fact that each constraint/cost is usually only relevant to part of the variables.
In this section, we use the notation ${\mathbb{NULL}}(\cdot)$ and ${\mathbb{ ROW}}(\cdot)$ to denote the null space and row space of the argument, respectively; use ${\rm{Null}}(\cdot)$ and ${\rm{ Row}}(\cdot)$ to denote a matrix whose \textit{rows} form an orthonormal basis of ${\mathbb{NULL}}(\cdot)$ and ${\mathbb{ ROW}}(\cdot)$, respectively. We use ${\rm rows}(\cdot)$ to denote the number of rows in the argument.

Before introducing our algorithm, we need to make some observations about the nature of the problem.
Since the feasible velocity $v$ under a HFVC may not be unique, the proper statement of the goal constraint is:
\begin{equation}
  \label{eq: interpretation of goal}
   Gv = b_G,\ \  \forall v \in \{v\ |\ Jv = 0, Cv=b_C\},
\end{equation}
i.e. we need to ensure all possible solutions satisfy the goal.
This is an inclusion relationship between the solution sets of two linear equations, which is equivalent to:
\begin{enumerate}
  \item The null space of $G$ contains the null space of $\left[\begin{array}{c}J\\C\end{array}\right]$;
  \item There exists a common special solution: $\exists v^*: Gv^* = b_G, Jv^* = 0, Cv^* = b_C$.
\end{enumerate}
We call them the \textit{Goal-Inclusion conditions} and will refer to them repeatedly. Condition 1) is equivalent to
\begin{equation}
  \label{eq: JC JG nullspace condition}
  {\mathbb{NULL}}(\left[\begin{array}{c}J\\C\end{array}\right]) \subseteq {\mathbb{NULL}}(\left[\begin{array}{c}J\\G\end{array}\right]),
\end{equation}
which further implies
\begin{equation}
  \label{eq: JC JG rank condition}
  {\rm rank}(\left[\begin{array}{c}J\\C\end{array}\right]) \ge {\rm rank}(\left[\begin{array}{c}J\\G\end{array}\right]).
\end{equation}
Due to the orthogonal complement relation between the row and null space of a matrix, the null space inclusion (\ref{eq: JC JG nullspace condition}) can be reformulated as a reverse row space inclusion:
\begin{equation}
  \label{eq: JCG rank condition}
  {\rm rank}\left(\left[\begin{array}{c}J\\C\end{array}\right]\right) = {\rm rank}\left(\left[\begin{array}{c}J\\C\\G\end{array}\right]\right).
\end{equation}

Our algorithm involves three steps. First, we derive the control axis directions to satisfy the Goal-Inclusion condition 1) while optimizing conditioning. Second, we compute the velocity control magnitudes to satisfy the Goal-Inclusion condition 2). Finally, we solve for the force control magnitudes to satisfy the guard conditions.

\subsection{Pick Control Axes to Optimize Conditioning} % (fold)
\label{sub:pick_force_controlled_directions}
The information of the control axes ($n_{av}, n_{af}$, and $T$) is contained in the velocity control coefficient matrix $C$~(\ref{eq:velocity control}): $C$ has $n_{av}$ rows; $C$ and its orthogonal complement forms $T$.

First thing we need to know about the velocity control is its dimension. We can compute the instantaneous directions that the system can move without conflicting the contact constraints by computing the null space $U$ of contact Jacobian:
\begin{equation}
  U = {\rm Null}(J).
\end{equation}
The last $n_a$ columns of $U$, i.e. the actuated part, indicate the directions in which the robot can move freely. It is a linear space, a basis of which can be computed as:
\begin{equation}
  \label{eq: U_bar}
  \bar U = {\rm Row}(US_a),
\end{equation}
where $S_a \in\mathbb{R}^{n\times n}$ is a selection matrix with only ones on the last $n_a$ diagonal entries. Any linear combinations of rows of $\bar U$ corresponds to a vector in ${\mathbb{NULL}}(J)$ and is thus a free robot motion direction. The dimensionality of $\bar U$ indicates the maximum dimension of velocity control we can apply:
\begin{equation}
  \label{eq: n_av upper bound}
  n_{av} \le {\rm rows}(\bar U)
\end{equation}
On the other hand, (\ref{eq: JC JG rank condition}) suggests the minimum dimension of velocity control required to satisfy the goal (\ref{eq:goal}):
\begin{equation}
  \label{eq: n_av lower bound}
  n_{av} \ge {\rm rank}(\left[\begin{array}{c}J\\G\end{array}\right])
  - {\rm rank}(J).
\end{equation}
Combining (\ref{eq: n_av upper bound}) and (\ref{eq: n_av lower bound}), we have a necessary condition for the feasibility of the problem:
\begin{equation}
  \label{eq:goal feasibility condition}
  {\rm rows}(\bar U) \ge {\rm rank}(\left[\begin{array}{c}J\\G\end{array}\right])
  - {\rm rank}(J).
\end{equation}
If equation~(\ref{eq:goal feasibility condition}) is not satisfied, the problem has an infeasible goal. Otherwise, we can choose the dimension of velocity control within (\ref{eq: n_av upper bound})-(\ref{eq: n_av lower bound}). Sometimes we want more velocity control so as to increase disturbance rejection ability \cite{Hou2020SharedGrasping}; sometimes we want less velocity control to have more compliance in the system \cite{HouICRA19Hybrid}. We solve both situations and leave this choice to the user.

If the maximal velocity control is needed, we can simply do velocity control in all directions in $\bar U$:
\begin{equation}
  \label{eq: max nav}
  n_{av} = {\rm rows}(\bar U),
\end{equation}
\begin{equation}
  \label{eq: max C}
  C = \bar U.
\end{equation}
Then we check equation~(\ref{eq: JCG rank condition}) to see if the problem is feasible.
Otherwise, if the minimal velocity control is desired, we take
\begin{equation}
  \label{eq: min nav}
  n_{av} = {\rm rank}(\left[\begin{array}{c}J\\G\end{array}\right])
  - {\rm rank}(J)
\end{equation}
Then the $n_{av}$ rows of velocity controls are linear combinations of rows of $\bar U$:
\begin{equation}
  \label{eq: C KU}
  C = K \bar U,
\end{equation}
where $K\in \mathbb{R}^{n_{av}\times {\rm rows}(\bar U)}$. We compute $K$ using the null space form of Goal-Inclusion condition 1), which implies
\begin{equation}
  C {\rm Null}(\left[\begin{array}{c}J\\G\end{array}\right]) = K \bar U {\rm Null}(\left[\begin{array}{c}J\\G\end{array}\right]) = 0.
\end{equation}
Then $K$ is an orthonormal basis of a null space:
\begin{equation}
  \label{eq: min K}
  K = {\rm Null}^T\left({\rm Null}^T(\left[\begin{array}{c}J\\G\end{array}\right])\bar U^T\right).
\end{equation}
The problem is feasible if $K$ has enough rows:
\begin{equation}
  \label{eq: min K condition}
  {\rm rows}(K) \ge n_{av}.
\end{equation}
If this is true, we keep the first $n_{av}$ rows of $K$ and recover $C$ from equation~(\ref{eq: C KU}). The $C$ obtained this way has orthonormal rows, since it is the product of two orthonormal matrices.

Note that equation~(\ref{eq: max C}) and (\ref{eq: C KU}) compute the velocity control direction $C$ in a closed form without explicitly optimize the crashing index~(\ref{eq:condition number}), however, they do find optimal solutions. As shown in our numerical experiments, equation~(\ref{eq: C KU}) always finds the solution with the minimal crashing index; equation~(\ref{eq: max C}) also always achieves the minimal crashing index among solutions with the same dimensionality.

After obtaining the velocity-controlled direction $C$, we compute the force-controlled direction as its orthogonal complement to make the velocity and force controls reciprocal.
Denote the last $n_a$ columns of $C$ as $R_{C}$, we can expand it into a full rank $R_a$:
\begin{equation}
  \label{eq:control dimensions}
  \begin{array}{rl}
  n_{af} &= n_a - n_{av}.
  \end{array}
\end{equation}
\begin{equation}
\label{eq:expand_R_a}
{R_a} = \left[ {\begin{array}{*{20}{c}}
{{\rm Null}{{({R_{C}})}^T}}\\
{{R_{C}}}
\end{array}} \right].
\end{equation}
Then we have $T=\rm{diag}(I_u, R_a)$.

We summarize the procedure in line~\ref{line:1} to line~\ref{line:v} in algorithm~\ref{alg:OCHS}. Note that the method avoids the non-convex optimization in \cite{HouICRA19Hybrid}.

\subsection{Solve for Velocity Control Magnitudes} % (fold)
\label{sub:solve_for_the_magnitude_of_velocity_control}
Next, we use Goal-Inclusion condition 2) to compute $b_C$. Compute a special solution $v^*$ from
\begin{equation}
  \label{eq: goal special solution}
  \left[\begin{array}{c}J\\G\end{array}\right]v^* = \left[\begin{array}{c}0\\b_G\end{array}\right]
\end{equation}
Such $v^*$ must exist, otherwise the goal itself is infeasible. Use it to compute the velocity control magnitude:
\begin{equation}
  \label{eq: compute velocity magnitude}
  w_{av} = b_C = Cv^*
\end{equation}
This choice of $b_C$ satisfies condition 2).

\begin{algorithm}[ht]
\caption{Optimally-Conditioned Hybrid Servoing}
\label{alg:OCHS}
\SetAlgoLined
\KwIn{Contact Jacobian $J$, Goal description $G, b_G$}
\KwIn{Guard condition,}
\KwOut{HFVC ($n_{af}, n_{av}, R_a, w_{av}, \eta_{af}$)}
\tcp{Solve for velocity control}
Compute free motion space $U$ under constraint\;\label{line:1}
Compute free robot motion space $\bar U$ (\ref{eq: U_bar}) \;
Check necessary feasibility condition (\ref{eq:goal feasibility condition}). \;
\uIf{Maximal velocity control dimension}{
  Take $\bar U$ as velocity-controlled directions (\ref{eq: max nav})(\ref{eq: max C}) \;
  Check goal feasibility using (\ref{eq: JCG rank condition})\;
  }
\uElse{
  Compute the minimal dimension of $C$ from (\ref{eq: min nav}) \;
  Solve for the coefficient matrix $K$ from (\ref{eq: min K}) \;
  Check goal feasibility using (\ref{eq: min K condition}) \;
  Compute the velocity control $C$ from (\ref{eq: C KU})\;
}
Complete control axes information using (\ref{eq:control dimensions})-(\ref{eq:expand_R_a}). \;
Compute the velocity control magnitude (\ref{eq: compute velocity magnitude}) using a special solution to (\ref{eq: goal special solution}). \; \label{line:v}
\tcp{Solve for Force control}
Compute the force control magnitude $\eta_{af}$ by solving the QP defined in Section~\ref{sub:solve_for_force_controlled_magnitudes}.\; \label{line:f}
\end{algorithm}

\subsection{Solve for Force Control Magnitudes} % (fold)
\label{sub:solve_for_force_controlled_magnitudes}
At this point we already know the dimensionality and directions of force control. The only remaining unknown variable in a HFVC is the value of the force control magnitude $\eta_{af}$. we find a solution by minimizing the magnitude of force variables:
\begin{equation}
  \label{eq: force cost}
  \min_{\lambda,\eta} \lambda^T\lambda + \eta_a^T\eta_a
\end{equation}
subject to the Newton's Second Law (\ref{eq:newton_second_law}) and the guard conditions (\ref{eq:guard_conditions}), which takes the form of a Quadratic Programming. This is a simplified version of the force magnitude computation in \cite{HouICRA19Hybrid} and it is slightly more efficient.

\subsection{Discussion} % (fold)
\label{sub:discussion}
An important advantage of this algorithm over the previous work \cite{HouICRA19Hybrid} is its ability to handle underactuated system.
\begin{wrapfigure}{r}{0.25\textwidth} %this figure will be at the right
    \centering
    \includegraphics[width=0.22\textwidth]{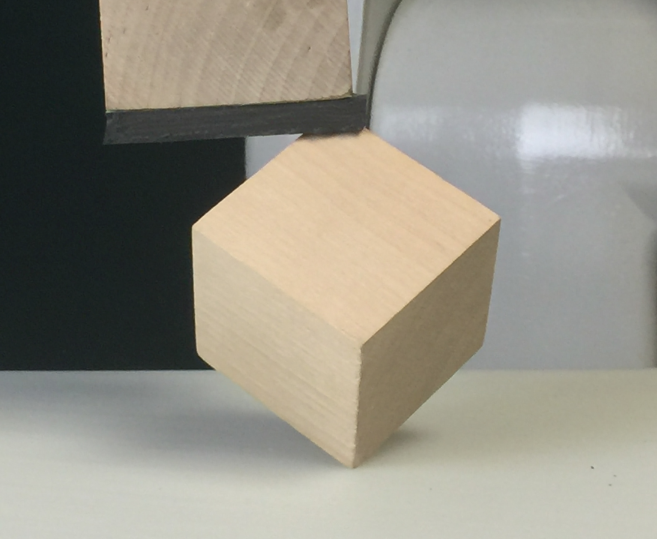}
    \caption{An underactuated example.}
    \label{fig:example_diamond}
\end{wrapfigure}
An example is a cube with one corner sticking on the ground and one corner sticking on the robot finger (Figure~\ref{fig:example_diamond}): the robot has no control over the rotation of the cube about the line between the two contact points. Still, a control problem on this object may still be feasible, e.g. if the goal is to lift the center of mass of the object. Condition (\ref{eq: JCG rank condition}) tells us whether this is the case or not.

% \subsection{Algorithm Complexity} % (fold)
% \label{sub:algorithm_complexity}
% The velocity control part
% QR: 2mn2
% O(mk^2)
% svd()

% U = null(J)'   n_c x n
% U_bar_n = orth(U_bar')';  n_av x n
% null_JG = null(JG);       n_c + n_goal x n

% k = null(K); ? n_av x n_av
% null(C_bar) n_av x na

% v_star = JG\b_JG

% !TEX root = ../ICRA21Analytical.tex

\section{Evaluation}
\label{sec:evaluation}
\subsection{Implementation} % (fold)
\label{sub:implementation}
In this section, we evaluate the performance of the OCHS algorithm in randomly generated manipulation problems. We implement OCHS in Matlab and test it on a desktop with an i7-9700k CPU clocked at 4.7GHz.
\subsection{Test Problems} % (fold)
\label{sub:test_problems}
We consider a rigid object with one to three environmental contacts and one to three rigid body fingers. Each rigid body has three DOFs in planar problems, or six DOFs in 3D problems. The settings are listed in Table~\ref{tb:test problems}, where `f' denotes a fixed (sticking) contact point, `s' denotes a sliding contact point. Each environment contact point can be sliding or sticking; finger contacts are all sticking.
\begin{table}[t]
\caption{Types of Randomly Generated Test Problems}
\label{tb:test problems}
\centering
\begin{tabular}{|l|l|l|l|l|}
\hline
\multirow{2}{*}{}       & \multicolumn{2}{l|}{Environment Contacts}                                                                             & \multicolumn{2}{l|}{Hand Contacts}                                            \\ \cline{2-5}
                        & \begin{tabular}[c]{@{}l@{}}\# Contact\\ Points\end{tabular} & \begin{tabular}[c]{@{}l@{}}Contact\\ Modes\end{tabular} & \begin{tabular}[c]{@{}l@{}}\# Contacts\\ per finger\end{tabular} & \# Fingers \\ \hline
\hline
\multirow{2}{*}{Planar} & 1                                                           & f,s                                                     & 1,2                                                              & 1          \\ \cline{2-5}
                        & 2                                                           & ss                                                      & 1,2                                                              & 1          \\ \hline
\multirow{3}{*}{3D}     & 1                                                           & f,s                                                     & 1,2,3                                                            & 1,2,3      \\ \cline{2-5}
                        & 2                                                           & ff,fs,ss                                                & 1,2,3                                                            & 1,2,3      \\ \cline{2-5}
                        & 3                                                           & ffs,fss,sss                                             & 1,2,3                                                            & 1,2,3      \\ \hline
\end{tabular}
\end{table}

\begin{table}[t]
\caption{Test results - Min Velocity Control Dimension}
\label{tb:test results1}
\begin{tabular}{|l|l|l|l|l|l|}
\hline
\multicolumn{2}{|l|}{\begin{tabular}[c]{@{}l@{}}Planar (6 DOF)\end{tabular}} & OCHS & OCHS(M) & HS3 & HS10 \\ \hline
\multirow{2}{*}{\begin{tabular}[c]{@{}l@{}}\# of\\ Problems\end{tabular}}       & Total    & \multicolumn{4}{c|}{6000}          \\ \cline{2-6}
                                                                                & Solved   & 5985 & 5981    &5974 & 5977 \\ \hline
\multicolumn{2}{|l|}{Average Crashing Index}                                               & 15.5 & 18.4    &19.5 & 19.9 \\ \hline
\multicolumn{2}{|l|}{ill-conditioned solutions}                                            & 15    & 19     & 26  & 23    \\ \hline
\multirow{2}{*}{\begin{tabular}[c]{@{}l@{}}Velocity\\ Time(ms)\end{tabular}}   & Average   & 0.14 & 0.14    & 1.74& 5.32 \\ \cline{2-6}
                                                                                & Worst    & 0.63 & 0.46    & 2.73& 7.40 \\ \hline
\multirow{2}{*}{\begin{tabular}[c]{@{}l@{}}Force\\ Time (ms)\end{tabular}}      & Average  & 0.99 & 0.94    & 1.28& 1.33 \\ \cline{2-6}
                                                                                & Worst    & 2.21 & 2.44    & 2.09& 3.34 \\ \hline
\end{tabular}
\vspace{3pt}

\begin{tabular}{|l|l|l|l|l|l|}
\hline
\multicolumn{2}{|l|}{\begin{tabular}[c]{@{}l@{}}3D, (12 to 24 DOF)\end{tabular}}    & OCHS & OCHS(M) & HS3 & HS10 \\ \hline
\multirow{2}{*}{\begin{tabular}[c]{@{}l@{}}\# of\\ Problems\end{tabular}}       & Total    & \multicolumn{4}{c|}{72000}          \\ \cline{2-6}
                                                                                & Solved   & 67506& 67399   & 65950& 65952 \\ \hline
\multicolumn{2}{|l|}{Average Crashing Index}                                               & 3.98 & 13.8    & 5.20 & 4.68  \\ \hline
\multicolumn{2}{|l|}{ill-conditioned solutions}                                            & 22   & 136     & 50   & 48  \\ \hline
\multirow{2}{*}{\begin{tabular}[c]{@{}l@{}}Velocity\\ Time(ms)\end{tabular}}   & Average   & 0.28 & 0.28    & 1.96 & 5.70  \\ \cline{2-6}
                                                                                & Worst    & 0.85 & 0.77    &3.82  &10.7 \\ \hline
\multirow{2}{*}{\begin{tabular}[c]{@{}l@{}}Force\\ Time (ms)\end{tabular}}      & Average  & 1.09 & 1.04    &1.57  &1.61 \\ \cline{2-6}
                                                                                & Worst    & 3.03 & 3.13    &12.3  &11.8 \\ \hline
\end{tabular}
\vspace{3pt}

\caption{Test results - Max Velocity Control Dimension}
\label{tb:test results2}
\begin{tabular}{|l|l|l|l|l|l|}
\hline
\multicolumn{2}{|l|}{\begin{tabular}[c]{@{}l@{}} \end{tabular}}        & OCHS  & OCHS(M) & HS3 & HS10 \\ \hline
\multirow{2}{*}{\begin{tabular}[c]{@{}l@{}}\# of\\ Problems\end{tabular}}       & Total    & \multicolumn{4}{c|}{30000}          \\ \cline{2-6}
                                                                                & Solved   & 27372 &27372    &22939 &22951 \\ \hline
\multicolumn{2}{|l|}{Average Crashing Index}                                               & 8.34  & 8.34    & 14.0 & 12.9 \\ \hline
\multicolumn{2}{|l|}{ill-conditioned solutions}                                            & 55    &55       & 37   & 31 \\ \hline
\multirow{2}{*}{\begin{tabular}[c]{@{}l@{}}Velocity\\ Time(ms)\end{tabular}}   & Average   & 0.17  &0.17     & 2.35 & 7.25 \\ \cline{2-6}
                                                                                & Worst    & 0.54  & 0.45    &3.99  & 28.39 \\ \hline
\multirow{2}{*}{\begin{tabular}[c]{@{}l@{}}Force\\ Time (ms)\end{tabular}}      & Average  & 1.04  & 0.98    &1.39  & 1.44 \\ \cline{2-6}
                                                                                & Worst    & 2.39  & 2.18    &3.76  & 3.62 \\ \hline
\end{tabular}
\vspace{3pt}

\end{table}

% \footnotetext{Only counting problems that are solved by all test algorithms. Unsolvable problems usually do not go through all steps and take less computation time. }
We randomly sample 1000 set of contact point locations and normals for each
contact mode setting, making a total of 6000 planar and 72,000 3D test problems.
The goal constraint is sampled randomly for each problem.
 % With a probability of 50\%, the first $n_u$ columns of the goal constraint coefficient matrix $G$ is set to zero, so the constraint only involves the robot.

The results are summarized in Table~\ref{tb:test results1} and \ref{tb:test results2},
the difference is that in Table~\ref{tb:test results2} we give the goal (\ref{eq:goal})
the maximum possible dimension, so all algorithms must select the maximum velocity
control dimension. OCHS is our algorithm with minimal velocity control, OCHS(M)
 denotes our algorithm with maximal velocity
control. As a comparison, we also show the performance of the original
hybrid servoing algorithm \cite{HouICRA19Hybrid}. HS3 uses three initial guesses and has a
good computation speed; HS10 uses ten initial guesses to search the problem excessively.
 % In both cases, we run projected gradient descent on
Both run each initial guess for 50 iterations.
\subsection{Results} % (fold)
\label{sub:results}
% The results are listed in Table~\ref{tb:test results}.
% The statistics for crashing index and computation time only counts problems that are solved by all four test methods.
%
% When the object is subject to two sticking contacts and
% one sliding contact with the environment in 3D, the object can not move. The randomly
% sampled sliding direction has a zero probability to align with the feasible object
% velocity under the other two sticking contacts. This is why each of the 3D tests
% has at least 1000 unsolvable problems. In 3D one finger tests, the object has
% uncontrollable DOFs when it has the following types of contact with the environment:
% 1. One sticking contact; 2. One sliding contact; 3. Two sliding contacts. The
% original hybrid servoing didn't handle this situation, so they have 3000 additional
% infeasible problems. However, the problem is indeed feasible when the goal
% constraint does not involve the uncontrollable object DOFs, which is why OCHS
% can solve more than 20k problems.
In all tests, OCHS consistently finds solutions with better crashing indexes than HS3 and HS10,
achieves lower average crashing index and fewer ill-conditioned solutions.
OCHS(M) has a larger crashing index in Table~\ref{tb:test results1} because it applies more dimensions of control.
When all algorithms selects the same dimension of velocity control, OCHS(M) also always achieves
better crashing indexes than HS3 and HS10 on every problem.

Both OCHS and OCHS(M) are notably faster than HS3 and HS10. The velocity part
of OCHS shows 7 to 13 times speedup comparing with HS3,
20 to 40 times speedup comparing with HS10. The force part of OCHS has a mild
speed up of 35\% due to a simpler problem formulation with less variables.

% !TEX root = ../ICRA21Analytical.tex
\section{EXPERIMENTS} % (fold)
\label{sec:experiments}
\begin{figure}[ht]
    \centering
    \includegraphics[width=0.45\textwidth]{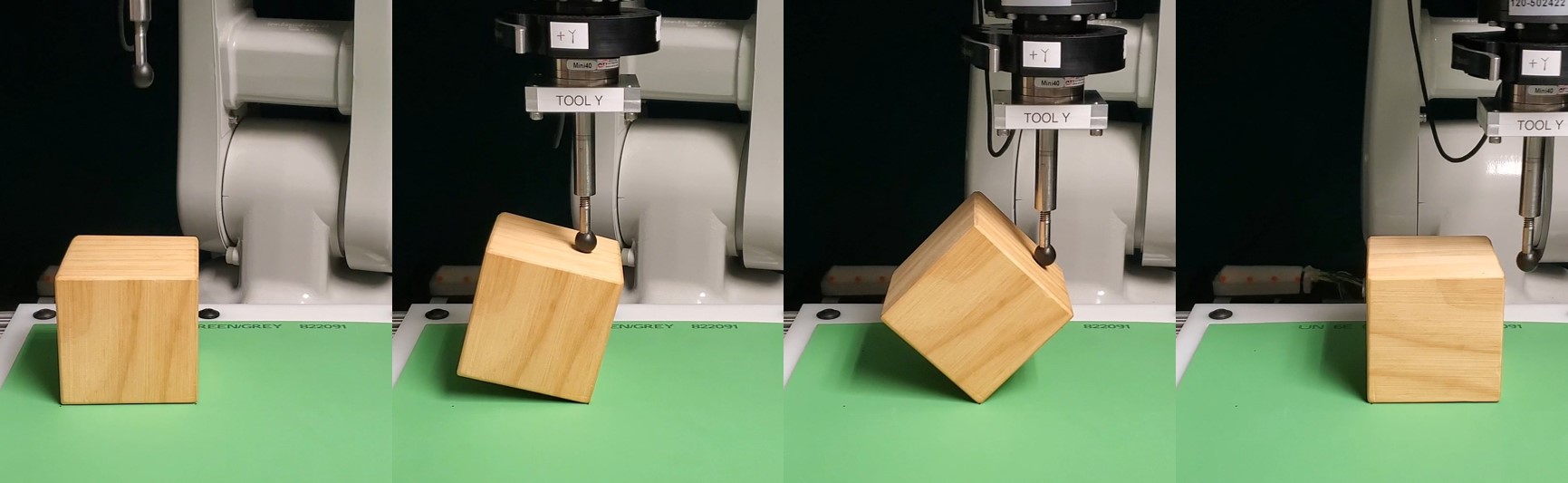}
    \caption{Robot flipping a block with one fingertip. The experiment runs 100 times with all successes.}
    \label{fig:block_tilting}
\end{figure}
We test the quality of our algorithm in several experiments. We implemented hybrid force-velocity control on a position-controlled ABB IRB 120 industrial robot with a wrist-mounted ATI Mini-40 force-torque sensor. In all experiments, we run OCHS off-line on a given motion plan to obtain a trajectory of HFVC, though the computation speed of OCHS supports feedback control at hundreds of Hz. In online execution, HFVC control loop is clocked at 200Hz. The lowest communication rate between the computer and the robot position controller is 250Hz with 25ms latency.

To compare with our previous work, we redo the block tilting experiment \cite{HouICRA19Hybrid} with the same setup, as shown in Figure~\ref{fig:block_tilting}. The block is a wooden cube with length 75mm. We place a 1.5mm rubber sheet on the table to increase friction. The robot hand is the same rubber ball as before. In our previous work, we perform block tilting 50 times and obtained 47 successes. With OCHS, we obtain a hundred consecutive successes at 50\% higher robot velocity \footnote{The full 40min video is available at \url{https://www.dropbox.com/s/ppimywwrgaelbw8/108_Block_tiltings.mp4?dl=0}}. The result demonstrate the ability of our algorithm to consistently produce solutions with good quality.

\begin{figure}[ht]
    \centering
    \includegraphics[width=0.45\textwidth]{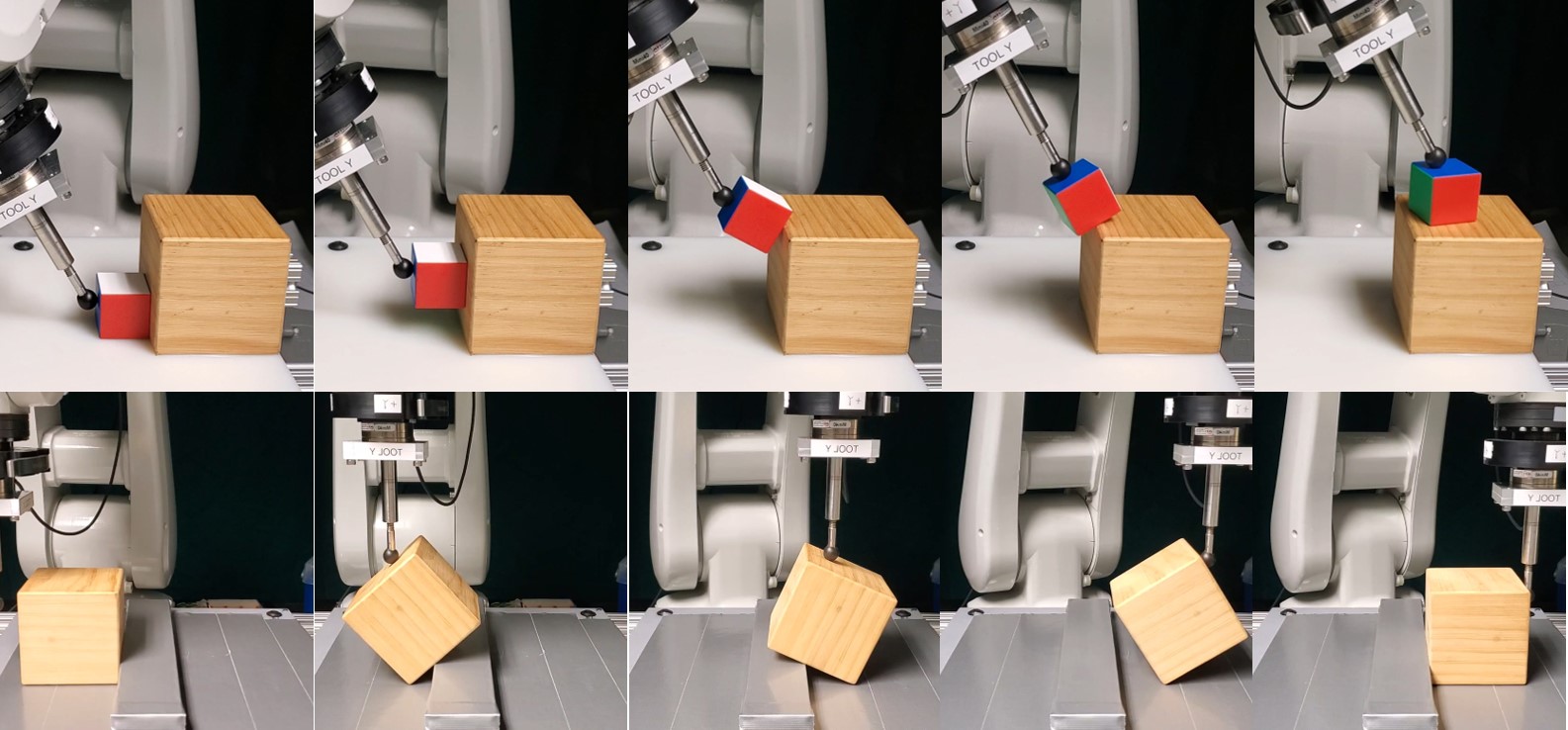}
    \caption{Two shared grasping tasks executed with OCHS. Top: moving an object up a stair. Bottom: transporting an object over an obstacle.}
    \label{fig:exp}
\end{figure}
OCHS can be used in shared grasping \cite{Hou2020SharedGrasping} to compute the control axes and velocity control. We use OCHS as a robust tracking controller to execute several shared grasping with motion plans computed using \cite{Cheng2020Contact}, as shown in Figure~\ref{fig:exp}.

% % \subsection{Resources} % (fold)
% % \label{sub:code}
% The Matlab implementation of the two algorithms along with several examples can be obtained from \href{https://github.com/yifan-hou/pub-icra19-hybrid-control}{https://github.com/yifan-hou/pub-icra19-hybrid-control}.
%  % You can also download our implementation of the low level hybrid force-velocity controller from \href{https://github.com/yifan-hou/forcecontrol}{another GitHub repository}.

% !TEX root = ../ICRA21Analytical.tex

\section{CONCLUSION} % (fold)
\label{sec:conclusion_and_discussion}

% There are a few corner cases where
% limitation
%     no enough constraint: all velocity control, no force control, can not satisfy guard condition

% Summary
In this work, we provide an algorithm to compute a hybrid force-velocity control
for manipulation under contact constraints. Our algorithm finds the solution that brings the best kinematic conditioning
of the manipulation system. We demonstrate that our algorithm reliably and quickly finds
the best solutions among all comparisons in extensive tests and experiments.

% future work
The algorithm itself can serve as a robust tracking controller to execute
a pre-computed motion plan. It is also a building block for the more comprehensive
contact stability analysis \cite{Hou2020SharedGrasping}.
Due to our algorithm's computational efficiency, it also has the potential to
be incorporated into a planning framework for robust manipulation planning.

\bibliographystyle{plain}
\bibliography{ICRA21Analytical}

\end{document}